\documentclass{article}
\usepackage{spconf,amsmath,graphicx}

\usepackage[
bookmarksopen,
bookmarksdepth=2,
colorlinks=true,
urlcolor=blue]{hyperref}

\title{Automatic Nonrigid Histological Image Registration with Adaptive Multistep Algorithm}
\name{Marek Wodzinski and Andrzej Skalski
\thanks{This work was funded by project no. 2018/29/N/ST6/00143 (NCN, Preludium 15).}}
\address{AGH University of Science and Technology \\ Department of Measurement and Electronics \\ Krakow, Poland \{wodzinski,skalski\}@agh.edu.pl}

\begin{document}
%\ninept
%
\maketitle
\begin{abstract}
In this paper, we present a short description of the method proposed to ANHIR challenge organized jointly with the IEEE ISBI 2019 conference. We propose a method consisting of preprocessing, initial alignment, nonrigid registration algorithms and a method to automatically choose the best result. The method turned out to be robust (99.792\% robustness) and accurate (0.38\% average median rTRE). The main drawback of the proposed method is relatively high computation time. However, this aspect can be easily improved by cleaning the code and proposing a GPU implementation.
\end{abstract}
\section{Introduction}

In this paper, we present our approach to solving the problem given in the ANHIR (Automatic Non-rigid Histological Image Registration) challenge organized jointly with the IEEE ISBI 2019 conference \cite{borovec}. We propose a robust, multimodal and fully automatic procedure which can successfully register even the most difficult registration pairs.

\section{Method}

\subsection{Overview}

During the algorithm development, we decided to not use one universal algorithm but several different approaches. Then, we proposed a method to automatically choose the best result. The procedure can be divided into five main steps:
\begin{enumerate}
\setlength{\itemsep}{0pt}
    \item Preprocessing.
    \item Initial registration (similarity/rigid transformation).
    \item Automatic decision to choose the best initial alignment.
    \item Nonrigid registration based on four different methods.
    \item Automatic decision to choose the best nonrigid registration.
\end{enumerate}

\subsection {Preprocessing}

The preprocessing consists of the following steps. Firstly, the data was converted to grayscale, smoothed and downsampled to lower resolutions. The resolution was different for the initial alignment and the nonrigid registration. The initial alignment was performed using the resolution where the maximum size was equal to 2048 pixels. For the nonrigid registration, the minimum size was equal to 1024 (locally affine registration dedicated to the missing data problem), 4096 (traditional, compositive Demons), 3000 (MIND, compositive Demons) and 4096 (TPS based on points from the initial alignment). 

After the downsampling, the image entropy was calculated for both images and the histogram of the image with lower entropy was matched to the histogram of the image with higher entropy. This step made it possible to use a traditional computer vision approach to calculate the initial similarity or rigid transformation. However, the MIND Demons used the images before the histogram matching since it was dedicated to purely multimodal images. The histogram matching for these cases was not improving the outcome and for several cases even led to worse results.

Finally, the images were zero-padded to the same size and the intensity was inverted.

\begin{figure*}[t]
	\begin{minipage}[b]{0.32\textwidth}
		\centering
		\centerline{\includegraphics[width=1\linewidth]{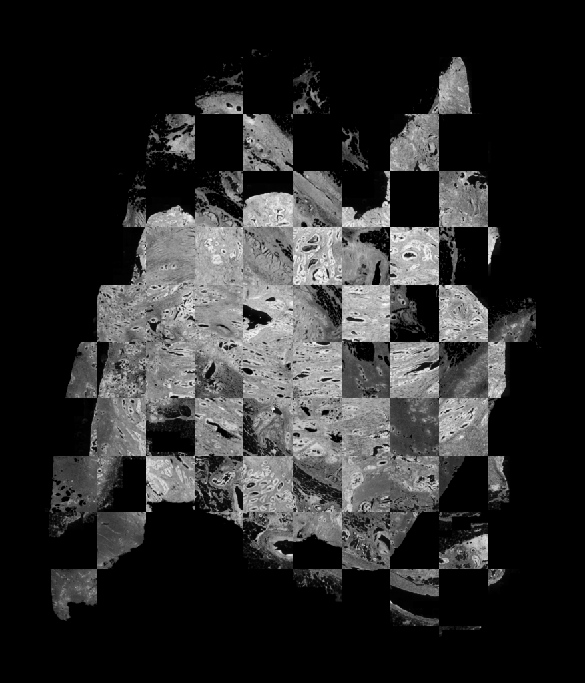}}
		\centerline{(a) Source/Target}\medskip
	\end{minipage}
	\hfill
	\begin{minipage}[b]{0.32\textwidth}
		\centering
		\centerline{\includegraphics[width=1\linewidth]{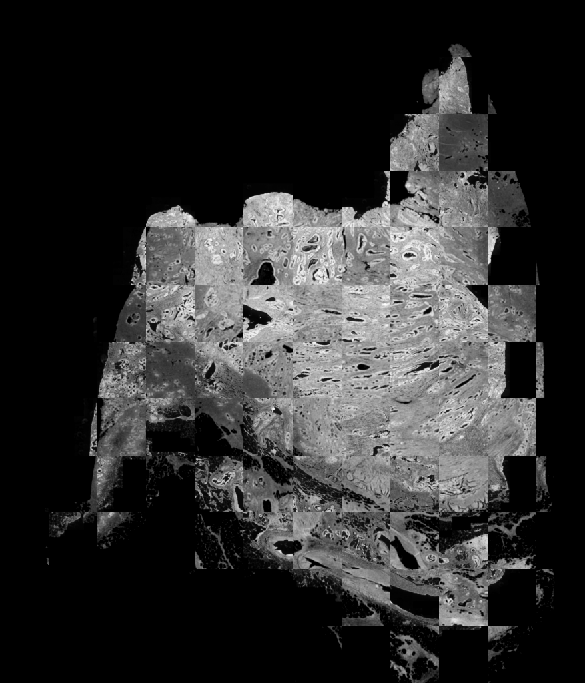}}
		\centerline{(b) IA Source/Target}\medskip
	\end{minipage}
	\hfill
	\begin{minipage}[b]{0.32\textwidth}
		\centering
		\centerline{\includegraphics[width=1\linewidth]{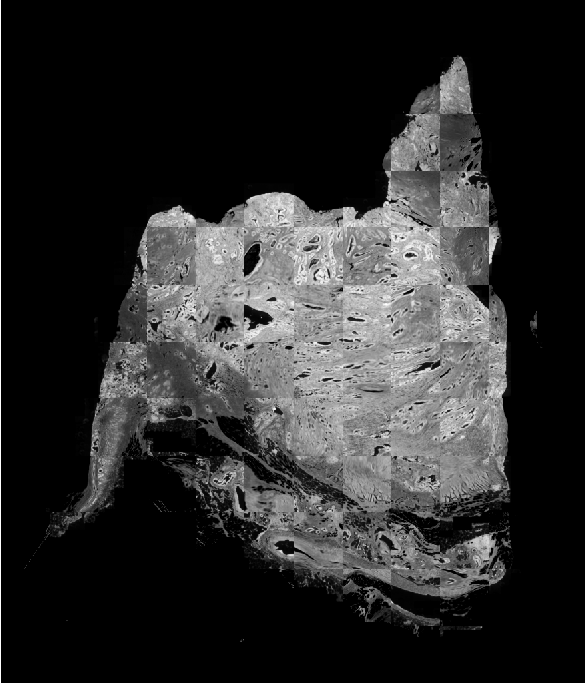}}
		\centerline{(c) IA + NR Source/Target}\medskip
	\end{minipage}
	
	\begin{minipage}[b]{0.32\textwidth}
		\centering
		\centerline{\includegraphics[width=1\linewidth]{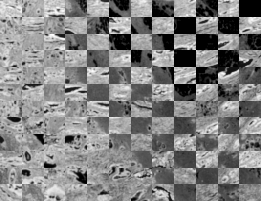}}
		\centerline{(a) Source/Target \textit{}}\medskip
	\end{minipage}
	\hfill
	\begin{minipage}[b]{0.32\textwidth}
		\centering
		\centerline{\includegraphics[width=1\linewidth]{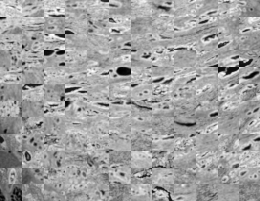}}
		\centerline{(b) IA Source/Target \textit{}}\medskip
	\end{minipage}
	\hfill
	\begin{minipage}[b]{0.32\textwidth}
		\centering
		\centerline{\includegraphics[width=1\linewidth]{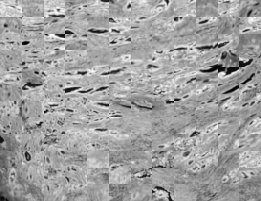}}
		\centerline{(c) IA + NR Source/Target}\medskip
	\end{minipage}

	\caption{An exemplary checkerboard visualization for image pair 137.}
	\label{fig:temp}
\end{figure*}

\subsection {Initial Alignment}

The initial alignment was based on rigid or similarity transformation. The initial alignment consisted of a two-step process. In the first step, three sets of features were calculated: SIFT \cite{SIFT}, ORB \cite{ORB}, SURF \cite{SURF}. Then, the RANSAC algorithm was used to calculate the similarity transformation for each determined feature pair. Finally, the evaluation of transformation quality was determined using the Dice coefficient. The Dice coefficient was calculated between thresholded \cite{LI} masks. Transformation with the highest coefficient value was selected. This procedure was mostly successful (more than 70\% of the pairs were initially aligned using this method). However, in case of fails (which were determined automatically), an alternative initial alignment procedure was proposed.

The alternative scenario used the binary version of source and target images, calculated using the Li thresholding \cite{LI}. Then, the centroid values were calculated for both the images and the desired rotation was optimized iteratively (with a given angle step). The Dice coefficient was used as a similarity measure. The output of this step was an input to a global affine registration based on SSD similarity metric with local intensity corrections. The correctness of this step was also automatically checked using the Dice coefficient between the thresholded source and target images.

The initial alignment with the automatic fail detector worked well for almost all 481 image pairs. All the fails were detected. For the pairs denoted as fails, the initial alignment was mostly not necessary. The average median rTRE for the training pairs after this step was close to 1.3\%.

 \begin{figure*}[t]
	\begin{minipage}[b]{0.755\textwidth}
		\centering
		\centerline{\includegraphics[width=1\linewidth]{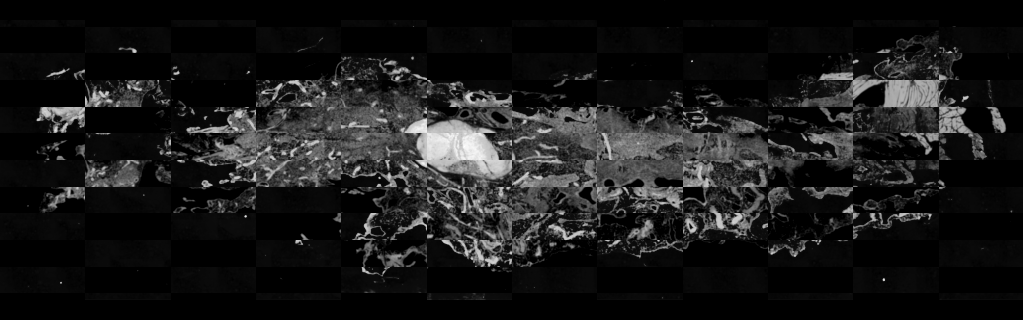}}
		\centerline{(a) Source/Target}\medskip
	\end{minipage}
	\hfill
	\begin{minipage}[b]{0.235\textwidth}
		\centering
		\centerline{\includegraphics[width=1\linewidth]{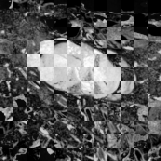}}
		\centerline{(a) Source/Target}\medskip
	\end{minipage}
	\hfill
	
	\begin{minipage}[b]{0.755\textwidth}
		\centering
		\centerline{\includegraphics[width=1\linewidth]{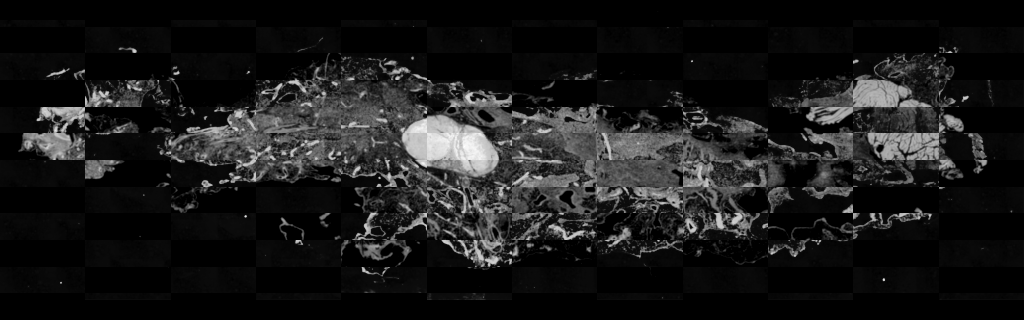}}
		\centerline{(b) IA Source/Target}\medskip
	\end{minipage}
	\hfill
	\begin{minipage}[b]{0.235\textwidth}
		\centering
		\centerline{\includegraphics[width=1\linewidth]{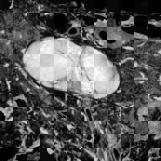}}
		\centerline{(b) IA Source/Target}\medskip
	\end{minipage}
	\hfill
	
	\begin{minipage}[b]{0.755\textwidth}
		\centering
		\centerline{\includegraphics[width=1\linewidth]{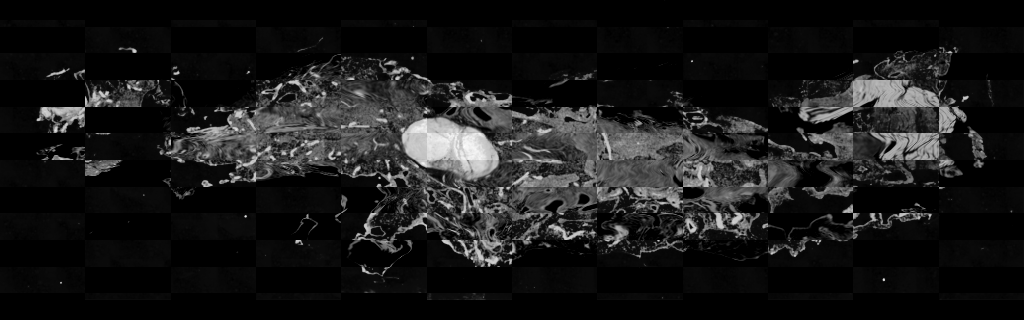}}
		\centerline{(c) IA + NR Source/Target}\medskip
	\end{minipage}
	\hfill
	\begin{minipage}[b]{0.235\textwidth}
		\centering
		\centerline{\includegraphics[width=1\linewidth]{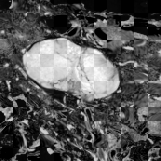}}
		\centerline{(c) IA + NR Source/Target}\medskip
	\end{minipage}
	\hfill

	\caption{An exemplary checkerboard visualization for image pair 426.}
	\label{fig:temp2}
\end{figure*}

\subsection {Nonrigid Registration}

The nonrigid registration used four different algorithms calculated in parallel: the local affine registration dedicated to the missing data problem \cite{periaswamy}, symmetric, compositive unimodal Demons \cite{thirion}, symmetric, compositive MIND Demons \cite{heinrich, reungamornrat} (but not diffeomorphic) and a simple TPS interpolation of best matches calculated during the initial alignment.

\subsubsection{Local Affine Registration}

We proposed the use of a nonrigid method based strongly on a method described in \cite{periaswamy}. This method was based on local affine transformation optimized jointly with a local brightness and contrast corrections. What is more, a probability that a given structure is missing in the second image was estimated in parallel, resulting in a multimodal algorithm resistant to the problem of missing data. The algorithm is described in details in \cite{periaswamy}. However, we introduced several modifications. Firstly, we calculated the contrast and brightness corrections during each iteration and composed them iteratively. This resulted in a much faster convergence compared to the original method. Secondly, we changed the algorithm to use the dense deformation field during its operation so it could be easily initialized from the initial registration, without introducing the interpolation error, as in the original implementation.

\subsubsection{Demons Registration}

The second applied nonrigid algorithm was the traditional compositive, symmetric Demons and the third method was also compositive, symmetric Demons but this time using SSD between MIND descriptors as the similarity metric. The reason to use this algorithm was an observation that the majority of the cases are extremely simple and an accurate, dense deformation field for these cases could improve the final rank significantly. The observation turned out to be indeed true and the traditional Demons (for cases where the histogram matching was accurate) and MIND Demons (for the remaining cases without the missing data) turned out to provide the best results.

\subsubsection{TPS Interpolation}

For some of the hardest cases, a relatively simple idea turned out to work really well (at least in terms of the median rTRE). We simply picked all the good matches from the initial alignment procedure (before the RANSAC algorithm) and used them as corresponding pairs for the thin plate splines registration. It turned out that the automatically detected feature key points were usually relatively close to the manually annotated landmarks and decreased the error significantly. The computation time of this algorithm was negligible compared to the remaining ones, it was just some kind of last resort which turned out to work well for some of the hardest cases, where other algorithms failed.

\subsubsection{Final Decision}

The final decision of the best nonrigid algorithm was based on the lowest MIND SSD between a masked content in the fixed image and the same mask in the transformed moving image. The mask was calculated the same way as described in the Initial Alignment section. The method detected all hard fails but sometimes failed to choose the best result. However, since the requirement was to make the method fully automatic, without any parameter tuning and human interaction, we decided that this small error was acceptable.

\section{Results}

The full quantitative results are available at: 
\begin{sloppypar}
\url{https://anhir.grand-challenge.org/evaluation/results/} (the AGH UST team).
\end{sloppypar} 

In short, we achieved an average median rTRE equal to 0.38\% with robustness equal to 99.792\% (a single fail). The average processing time according to the evaluation system was equal to 6.86 minutes, however, in reality, was a bit faster, using just a single core for the most time-consuming algorithm. The full and detailed results will be released soon by the challenge organizers. 

An exemplary visualization is shown in Figure \ref{fig:temp}. for a case we consider as "easy" and in Figure \ref{fig:temp2}. for the case we consider as "a bit harder".

\section{Discussion and Technical Details}

One could argue why all these algorithms were necessary. In fact, the Demons algorithm or the TPS interpolation were not mandatory to obtain really good, robust and useful results - the local affine registration was successful in almost all cases. In fact, the Demons usage decreased the average median rTRE by only 0.04\% and the thin plate splines corrected three hard fails to provide acceptable results (in term on the TRE). However, since the final challenge results were based on the rank of median rTRE, it was crucial to obtain as low rTRE as possible, even though it would not make any difference in the practical setup. In the extended version of this work we plan to propose only a single nonrigid algorithm but with the ability to operate on full image resolution in reasonable time.

Since the implementation of the locally affine registration dedicated to missing data problem was really poor (it was combined Python/Matlab implementation using only single core), the computation time was too high to use better image resolution. As a result, due to time limitations, we used quite a small resolution (1024 pixels for the larger dimension). Even though this resolution provided meaningful results, increasing the resolution was improving the rTRE. Since for majority of the registration pairs, the problem was trivial, using the optimized, multicore Demons and MIND Demons C++ implementation on much larger resolution (4096 and 3000 respectively, for the smaller dimension) improved the median rTRE a bit and increased the final rank. The computation time for the Demons and MIND Demons was negligible compared to the local affine transformation (few seconds for the Demons algorithms, over a dozen seconds for the MIND Demons), so we decided to simply use them in parallel and then automatically choose the best results. Of course, the Demons algorithm was not a good choice for the more difficult cases, however, the majority of the pairs were relatively easy problems.

In the future, we plan to reimplement the local affine registration dedicated to missing data problem using GPU (raw CUDA implementation or PyTorch - still to decide). Based on the initial results we think that it is possible to achieve processing time below one minute using the original resolution. As a result, the proposed method could be greatly simplified and the results could be much better.

\bibliographystyle{IEEEbib}
\bibliography{strings,refs}

\end{document}